\def\mode{icpm}
\def\icpm{%
\documentclass[conference]{./IEEEtran}		% ICPM
\bibliographystyle{./IEEEtran}
}
\def\article{%
\documentclass{article}	 % article
\setlength{\paperheight} {232.8mm}
\setlength{\paperwidth}  {151.5mm}
\setlength\voffset       {-23mm}
\setlength\hoffset       {-34mm}
\bibliographystyle{plain}
\usepackage[firstpage]{draftwatermark}
\SetWatermarkAngle{0}
\SetWatermarkFontSize{10pt}
\SetWatermarkHorCenter{153mm}
\SetWatermarkVerCenter{247mm}
\SetWatermarkLightness{0.6}
\SetWatermarkText{Draft{,} {\monthyeardate\today}}
}
\csname\mode\endcsname
\newif\ifshowtodos
\showtodosfalse	% do not show todo items

\newcommand{\authorartem}		 {Artem~Polyvyanyy}

\newcommand{\articletitle}   {Entropia: A Family of Entropy-Based Conformance\\ Checking Measures for Process Mining}
\newcommand{\articlesubject}  {Computer Science, Process Mining, Process Querying, Information Systems}
\newcommand{\articleauthors} {\authorartem}

\usepackage[usenames,dvipsnames]{xcolor}
\usepackage[algoruled,vlined,linesnumbered]{algorithm2e}
\usepackage{amsmath}
\usepackage{amsfonts}
\usepackage{mdframed}
\usepackage{paralist}
\setdefaultleftmargin{1em}{1em}{1em}{1em}{1em}{1em}
\usepackage{datetime}
\newdateformat{monthyeardate}{\monthname[\THEMONTH] \THEYEAR}
\usepackage[pdfsubject={\articlesubject},
                            pdfauthor={\articleauthors},
                            pdfkeywords={\articlesubject},
                            colorlinks=false,
                            pdfborder={0 0 0}]{hyperref}
\hypersetup{pdftitle={\articletitle}} % @update: Claudio Di Ciccio
\ifshowtodos
\usepackage[textsize=scriptsize,colorinlistoftodos]{todonotes}
\else
\usepackage[textsize=scriptsize,colorinlistoftodos,disable]{todonotes} % disable all todo notes
\fi
\let\todonote\todo
\renewcommand{\todo}[2]{\todonote[inline,color=red!20]{TODO (#1): #2}}
\newcommand{\done}[2]{\todonote[inline,color=green!20]{DONE (#1): #2}}

\usepackage{tikz,pgf,xcolor}
\usetikzlibrary{arrows,automata,trees,plotmarks,shadows,shapes}
\usepackage{pgfplots}
\usetikzlibrary{pgfplots.groupplots}
\usepackage{url}
\usepackage{mathtools} 
\usepackage[capitalise,nameinlink]{cleveref}
\usepackage{nicefrac}
\definecolor{mybluecolor}{RGB}{50,106,218}
\definecolor{myredcolor}{RGB}{176,53,53}
\definecolor{mygreencolor}{RGB}{93,172,0}
\definecolor{myyellowcolor}{RGB}{255,163,34}
\definecolor{mypurplecolor}{RGB}{86,35,132}
\definecolor{mytealcolor}{RGB}{30,161,165}

\newcommand{\splitatcommas}[1]{%
  \begingroup
  \ifnum\mathcode`,="8000
  \else
    \begingroup\lccode`~=`, \lowercase{\endgroup
      \edef~{\mathchar\the\mathcode`, \penalty0 \noexpand\hspace{-1pt plus 3em}}%
    }\mathcode`,="8000
  \fi
  #1%
  \endgroup
}

 \newcommand{\mset}[1] {\ensuremath [\splitatcommas{#1}]}
\newcommand{\msetel}[2]{{\ensuremath {{#1}^{#2}}}}

\newcommand{\fig}[9]{\begin{figure}[#1]
\vspace{#2mm}
\begin{center}
	\includegraphics[scale=#3,trim=#4]{#5}
\end{center}
\vspace{#6mm}
\caption{#7.}
\vspace{#8mm}
\label{#9}
\end{figure}}

\newcommand{\orcidartem}		{\href{https://orcid.org/0000-0002-7672-1643}{\protect\includegraphics[scale=0.05]{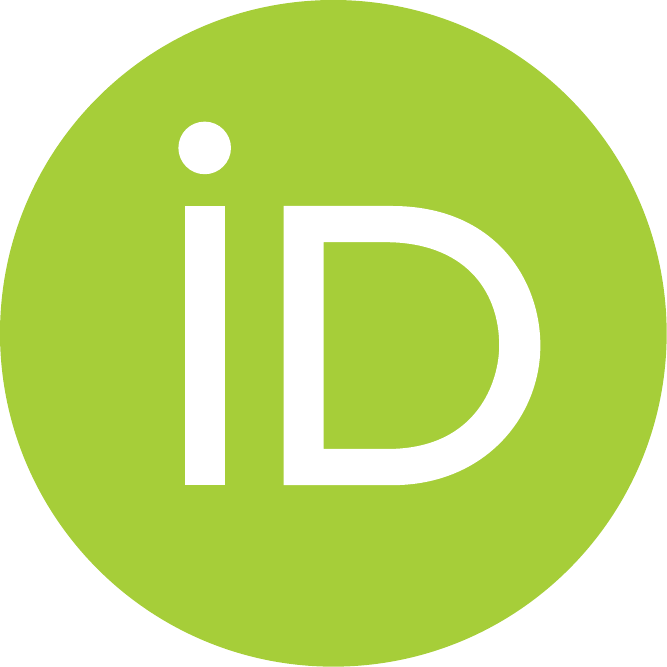}}}	% Artem Polyvyanyy
\newcommand{\orcidalistair}	{\href{https://orcid.org/0000-0002-6638-0232}{\protect\includegraphics[scale=0.05]{fig/orcid}}}	% Alistair Moffat
\newcommand{\orcidluciano}	{\href{https://orcid.org/0000-0001-9076-903X}{\protect\includegraphics[scale=0.05]{fig/orcid}}}	% Luciano Garcia-Banuelos
\newcommand{\orcidmatthias}	{\href{https://orcid.org/0000-0003-3325-7227}{\protect\includegraphics[scale=0.05]{fig/orcid}}}	% Matthias Weidlich
\newcommand{\orcidanna}			{\href{https://orcid.org/0000-0002-5088-7602}{\protect\includegraphics[scale=0.05]{fig/orcid}}} % Anna Kalenkova
\newcommand{\orcidjan}			{\href{https://orcid.org/0000-0002-7260-524X}{\protect\includegraphics[scale=0.05]{fig/orcid}}} % Jan Mendling
\newcommand{\orcidclaudio}	{\href{https://orcid.org/0000-0001-5570-0475}{\protect\includegraphics[scale=0.05]{fig/orcid}}} % Claudio Di Ciccio
\newcommand{\orcidsander}		{\href{https://orcid.org/0000-0002-5201-7125}{\protect\includegraphics[scale=0.05]{fig/orcid}}} % Sander Leemans
\newcommand{\orcidhanan}		{\href{https://orcid.org/0000-0001-5761-1345}{\protect\includegraphics[scale=0.05]{fig/orcid}}} % Sander Leemans

\newcommand{\ie}					{{i.e.,}}
\newcommand{\eg}					{{e.g.,}}

\newtheorem{mytheorem}		{Theorem}
\newtheorem{mydefinition}	{Definition}
\newtheorem{mylemma}			{Lemma}
\newtheorem{myproposition}{Proposition}
\newtheorem{mycorollary}	{Corollary}
\newtheorem{myexample}		{Example}
\newtheorem{myconjecture}	{Conjecture}

\newtheorem{myinvariant}	{Invariant}

\numberwithin{mytheorem}		{section}
\numberwithin{mydefinition}	{section}
\numberwithin{mylemma}			{section}
\numberwithin{myproposition}{section}
\numberwithin{mycorollary}	{section}
\numberwithin{myexample}		{section}
\numberwithin{myconjecture}	{section}
\numberwithin{myremark}			{section}
\numberwithin{myinvariant}	{section}

\addtocounter{mytheorem}{0}

\title{\articletitle}

\date{\today}

\author{
\IEEEauthorblockN{Artem~Polyvyanyy~\orcidartem}
\IEEEauthorblockA{
The University of Melbourne\\
\makebox[6cm][c]{artem.polyvyanyy@unimelb.edu.au}}
\and
\IEEEauthorblockN{Hanan~Alkhammash~\orcidhanan}
\IEEEauthorblockA{
The University of Melbourne\\
\makebox[6cm][c]{halkhammash@student.unimelb.edu.au}}
\and
\IEEEauthorblockN{Claudio~Di~Ciccio~\orcidclaudio}
\IEEEauthorblockA{
Sapienza Universit\`a di Roma\\
\makebox[6cm][c]{claudio.diciccio@uniroma1.it}}
\and
\IEEEauthorblockN{Luciano~Garc{\'{\i}}a{-}Ba{\~{n}}uelos~\orcidluciano}
\IEEEauthorblockA{
Tecnol\'ogico de Monterrey\\
\makebox[6cm][c]{luciano.garcia@tec.mx}}
\and
\IEEEauthorblockN{Anna~Kalenkova~\orcidanna}
\IEEEauthorblockA{
The University of Melbourne\\
\makebox[6cm][c]{anna.kalenkova@unimelb.edu.au}}
\and
\IEEEauthorblockN{Sander~J.~J.~Leemans~\orcidsander}
\IEEEauthorblockA{
Queensland University of Technology\\
\makebox[6cm][c]{s.leemans@qut.edu.au}}
\and
\IEEEauthorblockN{Jan~Mendling~\orcidjan}
\IEEEauthorblockA{
Wirtschaftsuniversit\"at Wien\\
\makebox[6cm][c]{jan.mendling@wu.ac.at}}
\and
\IEEEauthorblockN{Alistair~Moffat~\orcidalistair}
\IEEEauthorblockA{
The University of Melbourne\\
\makebox[6cm][c]{ammoffat@unimelb.edu.au}}
\and
\IEEEauthorblockN{Matthias~Weidlich~\orcidmatthias}
\IEEEauthorblockA{
Humboldt-Universit\"at zu Berlin\\
\makebox[6cm][c]{matthias.weidlich@hu-berlin.de}}
}

\newcommand{\ToolName}	{\textsl{Entropia}}

\begin{document}

\maketitle

%%%%%%%%%%%%%%%%%%%%%%%%%%%%%%%%%%%%%%%%%%%%%%%%%%%%%%%%%%%%%%%%%%%%%%%%%%%%%%%%
\begin{abstract}
\done{Artem}{Abstract}
This paper presents a command-line tool, called \ToolName, that implements a family of conformance checking measures for process mining founded on the notion of entropy from information theory.
The measures allow quantifying classical non-deterministic and stochastic \emph{precision} and \emph{recall} quality criteria for process models automatically discovered from traces executed by IT-systems and recorded in their event logs.
A process model has ``good'' precision with respect to the log it was discovered from if it does not encode many traces that are not part of the log, and has ``good'' recall if it encodes most of the traces from the log.
By definition, the measures possess useful properties and can often be computed quickly.
\end{abstract}
%%%%%%%%%%%%%%%%%%%%%%%%%%%%%%%%%%%%%%%%%%%%%%%%%%%%%%%%%%%%%%%%%%%%%%%%%%%%%%%%

%%%%%%%%%%%%%%%%%%%%%%%%%%%%%%%%%%%%%%%%%%%%%%%%%%%%%%%%%%%%%%%%%%%%%%%%%%%%%%%%
\section{Introduction}
\label{sec:1}
%%%%%%%%%%%%%%%%%%%%%%%%%%%%%%%%%%%%%%%%%%%%%%%%%%%%%%%%%%%%%%%%%%%%%%%%%%%%%%%%

\noindent
Process mining is a research field concerned with extracting knowledge from event sequence data that is stored in event logs. Conceptually, process mining techniques assume that events have at least three attributes: a timestamp, a case identifier and an activity type~\cite{Aalst2016}. Process mining techniques support various process analysis tasks including automatic process discovery, conformance checking, and variant analysis~\cite{Dumas2018}.

Conformance checking refers to those process mining techniques that compare the behavior captured in an event log with a normative process model~\cite{Carmona2018}. A key challenge for research on conformance checking is the definition of appropriate measures that quantify the extent of correspondence between the log and the model. A rich spectrum of measures have been proposed, albeit many of them in an ad hoc manner \cite{Tax2018}. The recent stream of work on entropy-based techniques provides a solid theoretical foundation for conformance checking measures with sound properties~\cite{Polyvyanyy2020TOSEM,PolyvyanyyK2019,KalenkovaP2020,Leemans2020,PolyvyanyyMG2020}, but in the past tool support has been somewhat limited.

In the paper at hand, we address this gap. 
Specifically, we present a command-line tool, called {\ToolName}, that implements entropy-based conformance checking techniques. 
The tool is publicly available\footnote{\url{https://github.com/jbpt/codebase/tree/master/jbpt-pm}} and supports process analysts in several scenarios in which commonalities and discrepancies between process models and event logs are measured.
Finally, the reader can take a look at a screencast\footnote{\url{https://youtu.be/RZVEFMuH684}} that demonstrates the tool and check the user guide\footnote{\url{https://github.com/jbpt/codebase/tree/master/jbpt-pm/entropia/guide.pdf}\label{footnote:tutorial}} that contains a comprehensive collection of examples and tutorials on using {\ToolName}.

The paper proceeds as follows. \Cref{sec:2} gives an overview of the theoretical foundations of conformance checking. \Cref{sec:3} introduces the {\ToolName} tool using a practical use case highlighting its analysis features. \Cref{sec:4} discusses the maturity of the work. \Cref{sec:5} provides illustrative examples. \Cref{sec:6} discusses computational performance and current limitations, before \Cref{sec:7} concludes.

%%%%%%%%%%%%%%%%%%%%%%%%%%%%%%%%%%%%%%%%%%%%%%%%%%%%%%%%%%%%%%%%%%%%%%%%%%%%%%%%
\section{Conformance Checking}
\label{sec:2}
%%%%%%%%%%%%%%%%%%%%%%%%%%%%%%%%%%%%%%%%%%%%%%%%%%%%%%%%%%%%%%%%%%%%%%%%%%%%%%%%

\noindent
The assessment of the model quality with respect to an event log is paramount for process mining~\cite{Aalst2016}.
Buijs et al.~\cite{Buijs.etal/IJCIS2014:QualityDimensionsinProcessDiscovery} introduce four main quality dimensions, namely \emph{fitness}, \emph{precision}, \emph{generalization}, and \emph{simplicity}, which are currently considered the de facto standard.
Fitness captures the degree to which the traces recorded in the event log can be replayed on the process model.
Precision penalizes the extra behavior introduced by the model that is not recorded in the event log.
Conversely, generalization indicates how well the model can support unforeseen traces.
Finally, simplicity denotes the capability of the model to express the behavior of the event log while keeping the model easy to understand.

Conformance checking techniques provide a number of approaches for assessing the four quality dimensions.
One can broadly classify them into two categories: \emph{descriptive} and \emph{quantitative}.
Descriptive techniques construct comprehensive artifacts that aim to explain various aspects of the studied criterion, {\eg} a description of all the commonalities and discrepancies between a trace and a process model.
Quantitative techniques measure the quantity of the studied phenomenon, {\eg} as a number between zero and one. 
Orthogonal to this classification is the partitioning of conformance checking techniques into non-stochastic and stochastic ones.
Stochastic conformance checking techniques study \emph{relations} between some stochastic aspects of the compared model and log, {\eg} distributions of traces recorded in the log and described in the model. 
In contrast, non-stochastic techniques, even though they may rely on the probabilistic aspects of the individual compared artifacts, do not analyze the relations between them.

\newdateformat{monthdayyeardate}{%
  \monthname[\THEMONTH]~\THEDAY, \THEYEAR}

%%%%%%%%%%%%%%%%%%%%%%%%%%%%%%%%%%%%%%%%%%%%%%%%%%%%%%%%%%%%%%%%%%%%%%%%%%%%%%%%
\section{Entropia}
\label{sec:3}
%%%%%%%%%%%%%%%%%%%%%%%%%%%%%%%%%%%%%%%%%%%%%%%%%%%%%%%%%%%%%%%%%%%%%%%%%%%%%%%%

\noindent
This section presents {\ToolName} by specifying the use cases it supports (\Cref{sec:3:use:cases}), the core principle behind the entropy-based measuring of precision and recall (\Cref{sec:3:entropy:cc}), and the command-line interface (CLI) of the tool (\Cref{sec:3:interface}).

%%%%%%%%%%%%%%%%%%%%%%%%%%%%%%%%%%%%%%%%%%%%%%%%%%%%%%%%%%%%%%%%%%%%%%%%%%%%%%%%
\subsection{Use Cases}
\label{sec:3:use:cases}
%%%%%%%%%%%%%%%%%%%%%%%%%%%%%%%%%%%%%%%%%%%%%%%%%%%%%%%%%%%%%%%%%%%%%%%%%%%%%%%%

\noindent
{\ToolName} implements the techniques for quantifying the \emph{precision} and \emph{recall} quality criteria in conformance checking presented in~\cite{Polyvyanyy2020TOSEM,PolyvyanyyK2019,KalenkovaP2020,Leemans2020}, and~\cite{PolyvyanyyMG2020}.
Two techniques~\cite{Leemans2020,PolyvyanyyMG2020} can be used to measure aspects that relate to stochastic precision and recall quality criteria.

%%%%%%%%%%%%%%%%%%%%%%%%%%%%%%%%%%%%%%%%%%%%%%%%%%%%%%%%%%%%%%%%%%%%%%%%%%%%%%%%
\subsection{Entropy-Based Conformance Checking}
\label{sec:3:entropy:cc}
%%%%%%%%%%%%%%%%%%%%%%%%%%%%%%%%%%%%%%%%%%%%%%%%%%%%%%%%%%%%%%%%%%%%%%%%%%%%%%%%

\noindent
The key idea for quantifying precision and recall between a model that describes ``relevant'' behavior and a model that captures ``retrieved'' behavior is to measure the magnitude of the behavior the two models share in relation to the magnitude of the behavior of one of the models.

Specific to the process mining context, one can think of an event log as a model that specifies the relevant behavior, {\ie} the behavior that provides information about the true behavior it was sampled from.
On the other hand, a process model discovered from an event log specifies the ``retrieved'' behavior, {\ie} the behavior the applied discovery algorithm constructed from the input event log.

Then, by following the principle for defining the corresponding quality criteria in information retrieval~\cite{Polyvyanyy2020TOSEM}, precision can be defined as the ratio of the magnitude of the shared behavior specified by the models of relevant and retrieved behaviors to the magnitude of the retrieved behavior.
Similarly, recall is the ratio of the magnitude of the shared behavior to the magnitude of the relevant behavior.

\Cref{fig:prec:and:rec} visualizes these ideas.
Note that $\mathit{rel} \cap \mathit{ret}$ refers to the behavior shared by the relevant and retrieved behaviors of the compared models.

\fig{h}{-3}{1.1}{0 0 0 0}{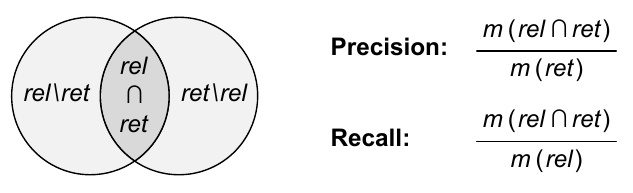}{-5}{Precision and recall quotients~\cite{Polyvyanyy2020TOSEM}}{0}{fig:prec:and:rec}

In the figure, \emph{rel} and \emph{ret} represent the relevant behavior and retrieved behavior, respectively.
Function $m$ is used to measure the magnitude of the corresponding (part) of the behavior.
The conformance checking approaches implemented in {\ToolName} interpret the behaviors of the compared models as collections of the traces that these models describe, where a trace is a sequence of process actions.
Function $m$ is implemented as the measure of the entropy of a collection of traces; note that the implemented conformance checking techniques use different notions of entropy, and in different ways, refer to \cref{sec:4}.

The benefit from using the entropy to measure the magnitudes of collections of traces when calculating precision and recall is twofold.
First, one can measure entropy of an arbitrary (potentially infinite) collection of traces.
Second, the entropy-based precision and recall measures can achieve a range of desired properties~\cite{Tax2018,Syring2019,Polyvyanyy2020TOSEM,Leemans2020}.

%%%%%%%%%%%%%%%%%%%%%%%%%%%%%%%%%%%%%%%%%%%%%%%%%%%%%%%%%%%%%%%%%%%%%%%%%%%%%%%%
\subsection{Interface}
\label{sec:3:interface}
%%%%%%%%%%%%%%%%%%%%%%%%%%%%%%%%%%%%%%%%%%%%%%%%%%%%%%%%%%%%%%%%%%%%%%%%%%%%%%%%

\noindent
As of August 2020, the {\ToolName} tool is in version 1.5. 
It is invoked by executing the command:

\vspace{-1mm}
{\footnotesize
\begin{verbatim}
    java -jar jbpt-pm-entropia-1.5.jar <options>
\end{verbatim}}
\vspace{-1mm}

\noindent
The core CLI options of {\ToolName} are listed in \cref{tab:entropia:cli:options}.

\begin{table}[htbp]
\scriptsize
  \centering
	\vspace{-3mm}
  \caption{\small Core CLI options of the {\ToolName} tool.}
	\vspace{-2mm}
\begin{tabular}{|l|c|c|l|}
  \hline
  \textbf{Option (full)}& \textbf{Option} & \textbf{Parameter} & \textbf{Description} \\
  \hline
  \hline
  \texttt{--help}			 & \texttt{-h}   & 									& print help message \\
  \hline
	\texttt{--relevant}	 & \texttt{-rel} &  \texttt{<path>} & model that describes relevant traces \\
  \hline
	\texttt{--retrieved} & \texttt{-ret} & \texttt{<path>}	& model that describes retrieved traces \\
  \hline
	\texttt{--silent} 	 & \texttt{-s}   &									& run tool in the silent mode \\
  \hline
  \texttt{--version}	 & \texttt{-v}   & 								  & get version of this tool \\
  \hline
\end{tabular}
\label{tab:entropia:cli:options}
\vspace{-1mm}
\end{table}

The \texttt{-h} and \texttt{-v} options print the help message and tool version, respectively, 
while options \texttt{-rel} and \texttt{-ret} are used to specify the models of relevant and retrieved traces, respectively.
To refer to a model, the user specifies its file path.
Option \texttt{-s} runs the tool in silent mode, in which
the result of the invocation is printed, without any debug information or execution data.
The tool accepts input models specified in one of the following formats: 
eXtensible Event Stream (XES)~\cite{7740858}, 
%Comma-Separated Values (CSV),
Petri Net Markup Language (PNML)~\cite{BillingtonCHKKPPSW03}, 
%Deterministic Finite Automaton (DFA),
Stochastic Petri Net Markup Language (sPNML),
Directly-Follows Graph (DFG), 
Stochastic Deterministic Finite Automaton (SDFA).
The latter three formats are specific to our tool.

Further CLI options allow selection of a conformance measure to be
applied to the input data, and configuration of it, and are detailed
in the next section.

%%%%%%%%%%%%%%%%%%%%%%%%%%%%%%%%%%%%%%%%%%%%%%%%%%%%%%%%%%%%%%%%%%%%%%%%%%%%%%%
\section{Maturity}
\label{sec:4}
%%%%%%%%%%%%%%%%%%%%%%%%%%%%%%%%%%%%%%%%%%%%%%%%%%%%%%%%%%%%%%%%%%%%%%%%%%%%%%%

\noindent
The work on the code base of the tool started in August 2017, together with the start of the work on the entropy-based approach for measuring precision and recall presented in~\cite{Polyvyanyy2020TOSEM}.
The tool is integrated into the jBPT library~\cite{Polyvyanyy2013a}, a compendium of open-source business process technologies, the work on which commenced in January 2009.

The approach presented in~\cite{Polyvyanyy2020TOSEM} suggests measuring precision and recall by interpreting the compared models, {\eg} process model and event log, as collections of traces that they describe.
The models are said to specify shared behavior if and only if they describe identical traces.
The magnitude of the behavior captured by each of the compared models, and of the behavior shared by the models, is determined as \emph{topological entropy}~\cite{Ceccherini-Silberstein2003} of the corresponding collection of traces.

In~\cite{PolyvyanyyK2019}, the authors generalize the approach from~\cite{Polyvyanyy2020TOSEM} by replacing every collection of traces involved in the calculations of the precision and recall measures with the collection of all subtraces of all the traces it contains.
Consequently, the shared behavior of two collections of traces is identified as a collection of all sequences of actions that are subtraces of some traces in both compared collections.
That is, this approach considers all the shared subsequences of actions in the compared models of traces for the measurements.

The measures described in~\cite{Polyvyanyy2020TOSEM} and~\cite{PolyvyanyyK2019} can be seen as extremes of the spectrum, with either \emph{no} or \emph{all} subtraces considered when determining the magnitudes of the collections of traces.
The approach presented in~\cite{KalenkovaP2020} allows for a flexible analysis.
In particular, based on knowledge of the compared models, the user can specify the maximal number of allowed skipped actions in a trace described by each of the models of traces for determining the shared subtraces.
This way, the user can tune the measures towards the desired sensitivity to the discrepancies in the compared behaviors.

\begin{table}[b]
\scriptsize
  \centering
	\vspace{-5mm}
  \caption{\small Specific CLI options of the {\ToolName} tool.}
	\vspace{-2mm}
\begin{tabular}{|c|c|l|c|}
  \hline
  \textbf{Option} & \textbf{Parameter} & \textbf{Description} & \textbf{Publ.} \\
  \hline
  \hline
  \texttt{-emp}			 & & exact matching precision & \cite{Polyvyanyy2020TOSEM} \\
  \hline
	\texttt{-emr}			 & & exact matching recall & \cite{Polyvyanyy2020TOSEM} \\
  \hline
	\texttt{-pmp}			 & & partial matching precision & \cite{PolyvyanyyK2019} \\
  \hline
	\texttt{-pmr}			 & & partial matching recall & \cite{PolyvyanyyK2019} \\
	\hline
	\texttt{-cpmp}		 & & controlled partial matching precision & \cite{KalenkovaP2020} \\
  \hline
	\texttt{-cpmr}		 & & controlled partial matching recall & \cite{KalenkovaP2020} \\
	\hline
	\texttt{-srel}		 & \texttt{<num>} & number of allowed skips in relevant traces & \cite{KalenkovaP2020} \\
  \hline
	\texttt{-sret}		 & \texttt{<num>} & number of allowed skips in retrieved traces & \cite{KalenkovaP2020} \\
	\hline
	\texttt{-sp}		   & & stochastic precision & \cite{Leemans2020} \\
	\hline
	\texttt{-sr}		   & & stochastic recall & \cite{Leemans2020} \\
  \hline
	\texttt{-r}			   & & entropic relevance & \cite{PolyvyanyyMG2020} \\
	\hline
\end{tabular}
\label{tab:entropia:cli:options:cont}
\vspace{0mm}
\end{table}

In~\cite{Leemans2020}, entropy is used to extend conformance checking to stochastic process mining.
An event log and a stochastic process model can be compared based on whether they exhibit the same control flow, but also based on whether the frequency of behavior in the event log matches the probabilities of behavior in the model.
To this end, both log and model are translated into stochastic deterministic finite automata, the conjunction of these automata is constructed, and the entropy of these three automata yields two measures: stochastic recall and stochastic precision.
In~\cite{Leemans2020}, an evaluation shows the practical applicability by searching for pairwise similar process models in a 4000-model repository.

Finally, the entropic relevance measure presented in~\cite{PolyvyanyyMG2020} is a stochastic conformance measure computed as the average number of bits required to compress ({\ie} to perform the entropy coding of) a trace from the log using the information on the relative likelihood of traces encoded in the model.

\Cref{tab:entropia:cli:options:cont} lists the tool options to select and configure the supported conformance measures.
\Cref{tab:entropy:techniques} then summarizes the characteristics of the conformance checking approaches implemented in {\ToolName} by specifying the input models of retrieved and relevant traces (L--event log, M--process model) supported by the approach presented in the corresponding publication (Publ.); the ability to address the stochastic aspect of the input models of traces (Stoch.); and the event log and process model formats supported (Log--event log, and Model--process model).

\begin{table}[t]
\scriptsize
  \centering
  \caption{\small Characteristics of conformance checking approaches.}
	\vspace{-2mm}
\begin{tabular}{|l|c|c|c|c|c|c|}
  \hline
  \!\textbf{Publ.}\! & \!\textbf{L-L}\! & \!\textbf{L-M}\! & \!\textbf{M-M}\! & \!\textbf{Stoch.}\! & \!\textbf{Log}\! & \!\textbf{Model}\! \\
  \hline
  \hline
	\cite{Polyvyanyy2020TOSEM}	& yes & yes & yes & no & XES & PNML \\
  \hline
	\cite{PolyvyanyyK2019}			& yes & yes & yes & no & XES & PNML \\
  \hline
	\cite{KalenkovaP2020}				& yes & yes & yes & no & XES & PNML \\
  \hline
	\cite{Leemans2020}					& yes & yes & yes & yes & XES & sPNML \\
  \hline
	\cite{PolyvyanyyMG2020}			& yes & yes & no & yes & XES & DFG, SDFA \\
  \hline
\end{tabular}
\label{tab:entropy:techniques}
\vspace{-3mm}
\end{table}

The approaches listed in \Cref{tab:entropy:techniques}, except the technique presented in~\cite{PolyvyanyyMG2020}, can be used to quantify precision and recall conformance criteria between two (possibly infinite) collections of traces.
The approach in~\cite{PolyvyanyyMG2020} measures the \emph{entropic relevance} of a stochastic process model to an event log.
Relevance reflects a compromise between the precision and recall criteria and has meaningful units.

%%%%%%%%%%%%%%%%%%%%%%%%%%%%%%%%%%%%%%%%%%%%%%%%%%%%%%%%%%%%%%%%%%%%%%%%%%%%%%%
\section{Examples}
\label{sec:5}
%%%%%%%%%%%%%%%%%%%%%%%%%%%%%%%%%%%%%%%%%%%%%%%%%%%%%%%%%%%%%%%%%%%%%%%%%%%%%%%

\begin{figure}[h!]
\begin{center}
\vspace{-3mm}
\begin{tikzpicture}[scale=0.65, transform shape, ->, >=stealth', shorten >=1pt, auto, node distance=20mm, on grid, semithick, transition/.style={fill=red!10, draw, minimum size=9mm, drop shadow}, place/.style={fill=blue!10, draw, circle, minimum size=9mm, drop shadow}]

\node[place,label=90:\small $p_0$]				(p0) 											{\Large $\bullet$};
\node[transition,label=90:\small $t_0$]		(t0) [right=of p0] 				{\Large $\texttt{a}$};
\node[place,label=90:\small $p1$] 				(p1) [above right=of t0]	{};
\node[place,label=270:\small $p_2$] 			(p2) [below right=of t0]	{};
\node[transition,label=90:\small $t_1$] 	(t1) [right=of p1]				{\Large $\texttt{b}$};
\node[transition,label=270:\small $t_3$] 	(t3) [right=of p2]				{\Large $\texttt{c}$};
\node[transition,label=0:\small $t_2$] 		(t2) [below right=of p1,xshift=17pt]				{\Large $\texttt{d}$};
\node[place,label=90:\small $p3$] 				(p3) [right=of t1]				{};
\node[place,label=270:\small $p4$] 				(p4) [right=of t3]				{};
\node[transition,label=90:\small $t_4$] 	(t4) [below right=of p3]	{\Large $\texttt{e}$};
\node[place,label=90:\small $p5$] 				(p5) [right=of t4]				{};

\path (p0) edge node {} (t0)
			(t0) edge node {} (p1)
					 edge node {} (p2)
			(p1) edge node {} (t1)
			(p2) edge node {} (t3)
			(t1) edge node {} (p3)
			(t3) edge node {} (p4)
			(t2) edge node {} (p1)
					 edge node {} (p2)
			(p3) edge node {} (t2)
					 edge node {} (t4)
			(p4) edge node {} (t2)
					 edge node {} (t4)
			(t4) edge node {} (p5)
;
\end{tikzpicture}
\vspace{-2mm}
\caption{A Petri net.}
\label{fig:petri:net}
\vspace{-2mm}
\end{center}
\end{figure}
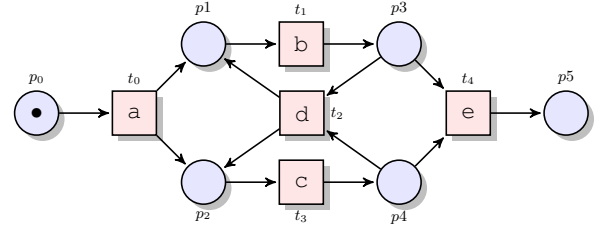

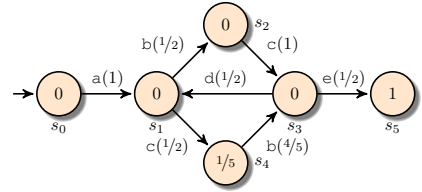
\begin{figure}[h!]
\begin{center}
\vspace{-5mm}
\begin{tikzpicture}[scale=0.65, transform shape, ->, >=stealth', shorten >=1pt, auto, initial text=, node distance=20mm, on grid, semithick, every state/.style={fill=orange!20, draw, circular drop shadow, text=black, minimum size=9mm}]
\node[initial,state,label=270:$s_0$]	(s0) 											{$0$};
\node[state,label=270:$s_1$]					(s1) [right=of s0] 				{$0$};
\node[state,label=0:$s_2$]						(s2) [above right=of s1]	{$0$};
\node[state,label=270:$s_3$] 					(s3) [below right=of s2]	{$0$};
\node[state,label=0:$s_4$] 					  (s4) [below right=of s1]	{$\nicefrac{1}{5}$};
\node[state,label=270:$s_5$] 				  (s5) [right=of s3]				{$1$};

\path (s0) edge node {$\texttt{a}(1)$} (s1)
			(s1) edge node {$\texttt{b}(\nicefrac{1}{2})$} (s2)
			     edge node[below, xshift=-14pt, yshift=-2pt] {$\texttt{c}(\nicefrac{1}{2})$} (s4)
			(s2) edge node {$\texttt{c}(1)$} (s3)
			(s3) edge node[above]  {$\texttt{d}(\nicefrac{1}{2})$} (s1)
					 edge node {$\texttt{e}(\nicefrac{1}{2})$} (s5)
			(s4) edge node[below, xshift=16pt, yshift=-2pt] {$\texttt{b}(\nicefrac{4}{5})$} (s3);
\end{tikzpicture}
\vspace{-2mm}
\caption{An SDFA.}
\label{fig:SDFA}
\vspace{-4mm}
\end{center}
\end{figure}

\noindent
In this section we provide some examples of {\ToolName},
using the Petri net $N$ in~\cref{fig:petri:net}, the SDFA $A$ in~\cref{fig:SDFA}, and an event log $E=\mset{\texttt{abce}, \texttt{ace}, \msetel{\texttt{bce}}{2}, \texttt{abcdcbe}, \texttt{abdcbe}, \texttt{aaacbe}}$.
Note that $E$ contains two instances of $\texttt{bce}$.

The entropy-based exact matching precision between $N$ and $E$ presented in~\cite{Polyvyanyy2020TOSEM} is computed using the CLI options:

\vspace{-1mm}
{\footnotesize
\begin{verbatim}
    -emp -rel=E.xes -ret=N.pnml
\end{verbatim}}
\vspace{-1mm}

To allow up to two skips in traces of the Petri net and up to one skip in the traces of the log, as described in~\cite{KalenkovaP2020}, when identifying similar traces in the computation of precision, these CLI options should be employed:

\vspace{-1mm}
{\footnotesize
\begin{verbatim}
    -cpmp -rel=E.xes -ret=N.pnml -srel=1 -sret=2
\end{verbatim}}
\vspace{-1mm}

\noindent
These options compute the entropic relevance of $A$ to $E$:

\vspace{-1mm}
{\footnotesize
\begin{verbatim}
    -r -rel=E.xes -ret=A.sdfa
\end{verbatim}}
\vspace{-1mm}

The computed values of exact matching precision, controlled partial matching precision, and entropic relevance using the above CLI options for net $N$, SDFA $A$ and log $E$ are 0.776, 0.833, and 11.368 bits, respectively.
Further examples of {\ToolName} and the serialized models and log used in the examples discussed above appear in the user guide.\textsuperscript{\ref{footnote:tutorial}}

%%%%%%%%%%%%%%%%%%%%%%%%%%%%%%%%%%%%%%%%%%%%%%%%%%%%%%%%%%%%%%%%%%%%%%%%%%%%%%%
\section{Discussion}
\label{sec:6}
%%%%%%%%%%%%%%%%%%%%%%%%%%%%%%%%%%%%%%%%%%%%%%%%%%%%%%%%%%%%%%%%%%%%%%%%%%%%%%%

\noindent
All the techniques implemented in {\ToolName} ver. 1.5 support process models that describe arbitrary (potentially infinite) collections of traces and impose no limitations on input logs provided that they are explicitly recorded and, thus, are finite.
However, process models must be bounded, {\ie} they must induce finite reachability graphs.
Various notions of semantic correctness of process models require process models to be bounded.
Nevertheless, process models used in practice can be incorrect, thus potentially unbounded.
Hence, each process model provided as input to {\ToolName}, which is not guaranteed by definition to be bounded, is tested by default for boundedness using LoLA ver. 2.0~\cite{Schmidt00}.
One can check if a process model is bounded using option \texttt{-b} of the tool.
If the boundedness of process models is established, one can invoke {\ToolName} with option \texttt{-t} to skip the model correctness tests.

{\ToolName} is implemented in Java and integrates with the LoLA tool compiled for Windows.
To use {\ToolName} on another platform, one needs to recompile LoLA for that platform.

Different conformance techniques implemented in {\ToolName} have different performance characteristics.
The computation time of entropic relevance~\cite{PolyvyanyyMG2020} is linear in the size of the event log (number of traces times average length of a trace). 
The computation of entropy-based precision and recall~\cite{Polyvyanyy2020TOSEM,PolyvyanyyK2019,KalenkovaP2020} is low polynomial in the size of the reachability graphs of the compared models of traces.
However, in practice, a reachability graph can be large, and possibly exponential in the size of the original model due to state explosion.
Empirical evidence suggests that the approach grounded in the exact matching of traces~\cite{Polyvyanyy2020TOSEM} runs in the order of seconds on real-world datasets, as the state explosion does not manifest often.
Grounding in the partial matching of traces~\cite{PolyvyanyyK2019}, on the other hand, induces large reachability graphs. Thus, it is recommended for small inputs, e.g., when calibrating a new automated process discovery technique.
The controlled partial matching technique~\cite{KalenkovaP2020} can
be configured by the user to the desired performance, balancing the
number of allowed mismatches between similar traces and runtime, with
fewer allowed mismatches allowing faster computation.
The techniques reported in~\cite{Polyvyanyy2020TOSEM,PolyvyanyyK2019}
constitute the two extremes of the trade-off spectrum.
Finally, the computation of the stochastic measures presented
in~\cite{Leemans2020} relies on an iterative procedure which
converges deterministically to the correct values, and is often
quick; but which can also lead to prolonged computation times on
real-world datasets.

Future work on {\ToolName} will both aim to improve the runtime
performance of the current techniques and also implement new
state-of-the-art information theoretic approaches to conformance
checking, including those that assess quality criteria beyond
precision and recall.
For example, in~\cite{KalenkovaPR2020} initial ideas are provided on
using entropy to measure the simplicity of a model automatically
discovered from a log.

%%%%%%%%%%%%%%%%%%%%%%%%%%%%%%%%%%%%%%%%%%%%%%%%%%%%%%%%%%%%%%%%%%%%%%%%%%%%%%%
\section{Conclusion}
\label{sec:7}
%%%%%%%%%%%%%%%%%%%%%%%%%%%%%%%%%%%%%%%%%%%%%%%%%%%%%%%%%%%%%%%%%%%%%%%%%%%%%%%

\noindent
This paper presents {\ToolName}, an open-source command-line tool for quantifying precision and recall conformance quality criteria in process mining.
The current version of the tool implements several measures, all grounded in the notion of the entropy of a collection of traces described by a process model or event log.
The supported measures can be used to assess both classical and stochastic precision and recall, and fulfill a wide range of desired properties suggested by the process mining community. 
The development of the tool's code base commenced in 2017 and is maintained by the authors of the implemented techniques.

\smallskip
\noindent
\textbf{Acknowledgment.}
Artem Polyvyanyy and Anna Kalenkova were in part supported by the Australian Research Council project DP180102839.

%%%%%%%%%%%%%%%%%%%%%%%%%%%%%%%%%%%%%%%%%%%%%%%%%%%%%%%%%%%%%%%%%%%%%%%%%%%%%%%%

\bibliography{bibliography}

\newpage
\listoftodos

\end{document}